\documentclass{article}
\usepackage{spconf,amsmath,graphicx}
\usepackage{cite}
\usepackage{amssymb,amsfonts}
\usepackage{algorithmic}
\usepackage{textcomp}
\usepackage{xcolor}
\usepackage{comment}
\usepackage{array}
\usepackage{multirow}

\usepackage{subcaption} 
\usepackage{enumitem} 
\usepackage{hyperref} 
\setlist{nosep, leftmargin=14pt}

\title{Weakly Supervised Cross-Modal Learning in High-Content Screening}

\name{\small
  Gabriel Watkinson\(^{1,2,3}\), 
  Ethan Cohen\(^{1,2,3}\), 
  Nicolas Bourriez\(^{2}\), 
  Ihab Bendidi\(^{2}\), 
  Guillaume Bollot\(^{3}\), 
  Auguste Genovesio\(^{2}\)
  \thanks{$^{1}$ Equal co-contribution. Thanks to IDRIS 2020-AD011011495 GENCI.}
}

\address{$^{2}$ IBENS, Ecole Normale Sup\'erieure, PSL University, Paris, France \\
$^{3}$ SYNSIGHT, Evry, France}

\begin{document}
\maketitle
\begin{abstract}

With the surge in available data from various modalities, there is a growing need to bridge the gap between different data types. 
In this work, we introduce a novel approach to learn cross-modal representations between image data and molecular representations for drug discovery. We propose EMM and IMM, two innovative loss functions built on top of CLIP that leverage weak supervision and cross sites replicates in High-Content Screening. 
Evaluating our model against known baseline on cross-modal retrieval, we show that our proposed approach allows to learn better representations and mitigate batch effect. In addition, we also present a preprocessing method for the JUMP-CP dataset that effectively reduce the required space from ~85Tb to a mere usable ~7Tb size, still retaining all perturbations and most of the information content. 

\end{abstract}


\section{Introduction}
Representation learning has become a cornerstone in many fields including drug discovery \cite{b1}. This approach is crucial for understanding drug mechanisms, predicting a drug’s activity and potential toxicity, and pinpointing chemical structures associated with diseases. However, there is a pressing challenge in capturing the complex relationship between the chemical structure of a molecule and its biological or physical outcomes. Current molecular representation methods focus mainly on encoding a molecule’s chemical properties, leading to unimodal representations \cite{b3}. The drawback here is that molecules, even if they have close structures, might behave very differently once in cells.

The advancement of high-content drug screens is pivotal in addressing these challenges \cite{b4}. Cell painting is a rich, image-based approach that captures the multifaceted cellular responses to chemical or genetic perturbations \cite{b8}. Analyses based on these images have been successful in tasks such as molecular bioactivity prediction \cite{b5} or mechanisms identification\cite{b7}.
The JUMP-CP Consortium \cite{b9} has compiled a significant repository of cellular images that results from treatment of 117K compound perturbations, making it the largest of its kind. However, the dataset's 100 Tb size is not its only daunting aspect. Compared to usual screens where replicates are often acquired in different plates in the same lab \cite{b8}, the JUMP-CP encompasses images and replicates acquired across multiple sites, processed in multiple batch and in different plates. While this experimental plan was carefully planned and made this dataset possible, it also introduces stronger substantial variability and batch effect \cite{b9}.

Additionally, drawing connections between molecular structures and high-content cellular images isn't straightforward. This task brings forward typical to cross-modal learning challenges. Firstly, it's not clear how molecules and images relate on a detailed level. Unlike simpler cross-modal data, like image-caption pairs, the features of a molecule are rarely directly visible in the images of treated cells. Secondly, inconsistencies and variations in the image data, due to noise and batch effects, are obstacles. Images from the same molecule might show significant differences, and traditional cross-modal pre-training methods might not handle this variability well, potentially affecting the model's ability to generalize.

In this work, we propose a solution to this problem. First, we bring a preprocessing approach providing an effective navigation to the JUMP-CP dataset, using only about 1/10th of the storage requirements. We then introduce two new objective functions, designed to leverage the subtle supervision present in high-content screening data. Given that each molecule treatment has multiple replicates and therefore corresponding images, our objective functions, named Extra Modality Multiview (EMM) loss and Intra Modality Multiview (IMM) loss, aim to strengthen the learnt relationship between molecular and phenotypic data while minimizing batch effects.

\section{Related Work}
\subsection{Multimodal Learning in Text and Computer Vision}

Multimodal learning, particularly in Natural Language Processing (NLP) and Computer Vision, has gained significant attention. The synergy between different data modalities offers a comprehensive understanding of the relationships between them. Seminal works like CLIP \cite{b10} have demonstrated the effectiveness of combining vision and text using contrastive learning. FLAVA \cite{b12} introduces a holistic approach to multimodal learning by targeting all modalities simultaneously. \cite{b13} presents a unique perspective by using multiple views of the same object, enhancing robustness across modalities.

\subsection{Machine Learning in Biological Research}

Machine Learning in drug discovery is foreseen to reduce drug development costs. Traditional methods relied on using precomputed molecular features, such as ECFP to help predict drug effect \cite{b14}.
However, with the rise of Deep Learning, Graph Neural Networks (GNN) have gained traction for encoding molecular structures to learn a useful representation. Variations of GNNs, like GCN \cite{b16}
and GIN \cite{b18}, have shown variations in performance based on the task at end. 
Given the high costs associated with molecular data acquisition, self-supervised learning methods such as Masked Graph Modelling \cite{b21} have gained prominence.

\subsection{Multimodal Learning in Drug Discovery}

Despite advancements in molecular representation learning, bridging the gap to the cellular phenotype response remain challenging. High-Content Screening \cite{b8} offers a wealth of data yet to be fully explored. Models such as MIGA \cite{b23}, CLOOME \cite{b24}, and MOCOP \cite{b25} have sought to harness this data. They employ contrastive learning to link molecular perturbations to microscopic images. While CLOOME uses fingerprints for molecular representations, MOCOP leverages the extensive JUMP CP dataset \cite{b9} but using precomputed handcrafted features. MIGA integrates both a traditional image encoder and a GNN, ensuring optimal representation power while also employing self-supervised learning techniques. However, none of the previous methods leverage the weak supervision offered by sample replicates in high-content screening data. 

\section{An Efficient Preprocessing of the JUMP-CP Dataset: } \label{JUMP} 

The JUMP CP dataset encompasses over 8 million high-resolution images. Each image contains at least 5 channels and has dimensions of around \(1000 \times 1000\) pixels encoded in 16-bit, culminating in a total dataset size surpassing 100 Tb. 
Each perturbation was replicated in different sources (i.e laboratories). To render this dataset more manageable:
\begin{itemize}
    \item 16-bit images were converted to 8-bit by rescaling the pixel values from the first and ninety ninth percentiles to 0 and 255 in order to handle image artefacts and gain a factor of 2.
    \item After testing various formats, we compressed all image from tiff to 5 single channel PNGs to gain a factor 1.9.
    \item For dimension reduction and consistency across the whole dataset, images underwent center cropping and resizing from \(1000 \times 1000\) to \(768 \times 768\) to gain a factor 1.7.
    \item A random selection of 6 views, out of 6 to 9 depending on the source, was performed for each treatment well, and 3 random views per control well (constituting one-sixth of the dataset) were curtailed to a factor of 1.8. Note that a careful selection of images had to be removed from one of the source (source 7) due to important protocol variations.
\end{itemize}

\noindent Through these transformations, the 84.7 Tb dataset was condensed to 7.4 Tb, ensuring the retention of pivotal information. The modified dataset maintained the 117k compounds, all the CRISPR and ORF perturbations, but scaled down the image count from 8 million to 4.4 million. The majority of conditions comprise approximately 30 images, as a result of the six fields of view per well and the five replicates per condition. 
Code for the dataset preprocessing is available \href{https://github.com/gwatkinson/jump_download}{here}.

\section{Methods}

In order to perform weak supervision, we formatted our dataset as a set of observations, where each entry associates a compound treatment with multiple images coming from different plates. 

\subsection{Image Encoder}

We used a ResNet34 as our image encoder. 
While being efficient, adopting this architecture necessitated modifications of the first convolution layer to cater to the 5-channel input and the high-resolution nature of our dataset images.

\subsection{Molecule Encoder}

For the molecule encoder, we harness the capabilities of Graph Neural Networks (GNNs) by using the variant known as Message Passing Neural Networks (MPNN). This encoder iteratively refines node representations by gathering information from neighboring nodes, culminating in a graph-level representation for compounds. We explored various architectures, including GIN and PNA \cite{b26}, and used the latest as it showcased faster and superior learning capabilities.

\begin{figure*}
  \includegraphics[width=\textwidth,height=3.8cm]{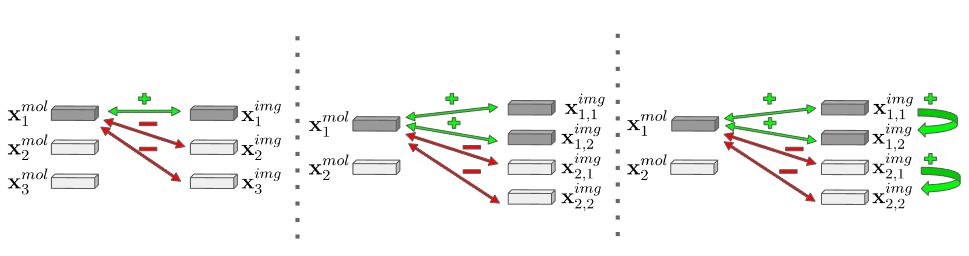}
  \caption{Depiction of the proposed methodologies. From left to right: the CLIP framework, followed by the EMM and IMM objectives. "+" denotes positive pairs while "-" denotes negative pairs.}
  \label{fig:all_frame}
\end{figure*}

\subsection{Training Objectives}

Inspired by multimodal text-image learning through multiple object views \cite{b13}, and contrastive pretraining of molecules using a variety of 3D configurations \cite{b28}, we defined a CLIP-like loss function that includes the multiple views available for each molecule, aiming to enhance the shared representation and mitigate the so-called batch effect. Furthermore, we designed two additional loss functions: the Extra Modality Multiview (EMM) and the  Intra Modality Multiview (IMM), see Figure \ref{fig:all_frame}. 

EMM focuses only on extra-modality pairings and establishes links between the molecule and the images on a one-to-one basis, without creating explicit connections between the images themselves. In practice, we sample \(N\) instances from the dataset 

\[
\left\{\mathbf{x}^{mol}_i, \left\{\mathbf{x}^{img}_j\right\}_{j \in E_i}\right\}_{i=1,\dots,N}
\]

For each compound \(\mathbf{x}^{mol}_i\), we select \(M\) images from \(E_i\), denoted by $\left(\mathbf{x}^{img}_{i, k}\right)_{k=1, \dots, M}$ preferably from different batches to ensure source diversity. The EMM loss is then expressed as:

\vspace{10pt} 
\begin{equation}
    \mathcal{L}_{EMM} = - \frac{1}{N} \sum_{i=1}^N \left[ \log \frac{
    \sum_{k=1}^M e^{\frac{\langle\mathbf{u}_i^{mol}, \mathbf{u}_{i, k}^{img}\rangle}{\tau}}
    }{
    \sum_{j \ne i}^N \sum_{k=1}^M e^{\frac{\langle\mathbf{u}_i^{mol}, \mathbf{u}_{j, k}^{img}\rangle}{\tau}}
    } \right]
\end{equation}

Furthermore, we explore the Intra Modality Multiview (IMM) loss, which considers inter-image links:

\vspace{10pt} 
\begin{equation}
    \mathcal{L}_{IMM} = \mathcal{L}_{EMM} - \frac{\gamma}{N} \sum_{i=1}^N \left[ \log \frac{
    \sum_{(a, b) \in B_M} e^{\frac{\langle\mathbf{u}_{i, a}^{img}, \mathbf{u}_{i, b}^{img}\rangle}{\tau}}
    }{
    \sum_{j \ne i}^N 
    \sum_{(a, b) \in B_M} e^{\frac{\langle\mathbf{u}_{i, a}^{img}, \mathbf{u}_{j, b}^{img}\rangle}{\tau}}
    } \right]
\end{equation}

where \(B_M\) is the set of all possible index pairs between 1 and \(M\) and $\gamma$ a parameter  (empirically set to 0.5). This loss ensures that images corresponding to the same perturbation yield analogous representations, thus potentially counterbalancing inter-source, inter-well and inter-image batch effects.

\section{Experiments}

\subsection{Evaluation Tasks}

We assessed our models with cross-modal retrieval and batch effect evaluation tasks. 

\textbf{Cross-Modal Retrieval: }
Our model aims to learn the relationship between molecules and their corresponding images of treated cells. It is pivotal for drug discovery to identifying similar compounds with dissimilar morphological effects or reversely. 
We evaluate these aspects using both image to molecule retrieval as well as molecule to image retrieval.  
To quantify this task, we use the following metrics: HitRate@1, HitRate@5, HitRate@5 and Mean Reciprocal Rank (MRR). 

\textbf{Batch Effect Evaluation: }
Inspired by \cite{b29}, we define a methodology to quantify the
batch effect on our trained image encoder. Employing various train/test splits for a given image task, we monitor accuracies on random, source-separated (NSS), batch-separated (NSB), and plate-separated (NSP) splits. Our task uses the TARGET2 plates from the JUMP CP dataset, that was unseen during contrastive pretraining. 
The goal of this task is to predict a compound's target gene from its image.
We used a subset of 9 targets with about 3k images each, employing a train/test split of 8:2 and repeating the classification five times. Using logistic regression and KNN on precomputed embeddings, our aim wasn't peak accuracy but to measure the loss of accuracy due to batch effects.

If we note $Acc_{Rand}, Acc_{NSS}, Acc_{NSB}, Acc_{NSP}$ the accuracy on the random, NSS, NSB and NSP splits respectively, we look at the following metric:
\begin{align}
    G_{NSS} &= \frac{Acc_{NSS}}{Acc_{Rand}} \tag{Source Generalisation} \\
    G_{NSB} &= \frac{Acc_{NSB}}{Acc_{Rand}} \tag{Batch Generalisation} \\
    G_{NSP} &= \frac{Acc_{NSP}}{Acc_{Rand}} \tag{Plate Generalisation}
\end{align}

\subsection{Experimental Setup}
We analyzed our model performance using the JUMP CP dataset, comprising 117k compounds from 13 sources. Using a 4k compounds validation set for monitoring and early stopping, and another distinct 4k compound set for retrieval tasks. We use AdamW optimizer with a weight decay of \(0.05\) and trained over 200 epochs with a cosine annealing learning rate scheduler and a 10-epoch linear warmup.
Training took three days for CLIP and seven days for IMM and EMM, on 3 NVIDIA V100 GPUs.

For training, we employed larger image sizes, particularly random crops to \(512\times 512\) pixels, supplemented by image resizing. Following \cite{b30} for biological image augmentations, we integrated flips and color changes at a \(0.4\) probability to mitigate overfitting. Our embeddings use a dimensionality of 1024 and a batch size of 384 for all models. Code for all models and training is available \href{https://github.com/gwatkinson/jump_models}{here}.

\section{Results}

Throughout this section, the various models are referenced as follows: "Random" for the random baseline, "CLIP" for the CLIP baseline, "EMM" for models trained using the EMM loss, and "IMM" for models trained using the IMM loss. All these models have otherwise the same encoders and experimental setups, differing only in their objective functions.

\subsection{Cross-modal retrieval}

Our initial evaluation focused on retrieval performances on the held-out set. As displayed in Table \ref{tab:1:100_img2mol} and Table \ref{tab:1:100_mol2image}, the EMM and IMM models significantly surpass the CLIP baseline in both retrieval tasks. 
While this enhancement suggests that integrating multiple views may refine retrieval capabilities, we nevertheless observed than the best framework depends on the task. This could be explained by the fact that IMM better captures the cell phenotype through the intra-image loss while EMM rather focuses on linking close molecules to multiple cell images by not explicitly matching these images.

\begin{table}[h]
    \centering
    \small
    \caption{Image to molecule retrieval results on the held out set, in the 1:100 configuration. The best and second-best values are \textbf{bolded} and \underline{underlined}, respectively.}
    \begin{tabular}{l|ccccc}
        \hline
        Experiment & HR@1 & HR@3 & HR@5 & HR@10 & MRR \\
        \hline
        Random & 0.010 & 0.030 & 0.050 & 0.100 & 0.052 \\
        CLIP & 0.077 & 0.192 & 0.284 & 0.452 & 0.191 \\
        EMM & \underline{0.085} & \underline{0.211} & \underline{0.295} & \underline{0.444} & \underline{0.198} \\
        IMM & \textbf{0.096} & \textbf{0.236} & \textbf{0.342} & \textbf{0.526} & \textbf{0.225} \\
        \hline
    \end{tabular}
    \label{tab:1:100_img2mol}
\end{table}

\begin{table}[h]
    \small
    \centering
    \caption{Molecule to image retrieval  on the held out set, in the 1:100 configuration. The best and second-best values are \textbf{bolded} and \underline{underlined}, respectively.}
    \begin{tabular}{l|ccccc}
        \hline
        Experiment & HR@1 & HR@3 & HR@5 & HR@10 & MRR \\
        \hline
        Random & 0.010 & 0.030 & 0.050 & 0.100 & 0.052 \\
        CLIP & 0.066 & 0.172 & 0.264 & 0.423 & 0.176 \\
        EMM & \textbf{0.103} & \textbf{0.238} & \textbf{0.335} & \textbf{0.500} & \textbf{0.225} \\
        IMM & \underline{0.077} & \underline{0.206 }& \underline{0.306} & \underline{0.498} & \underline{0.201} \\
        \hline
    \end{tabular}
    \label{tab:1:100_mol2image}
\end{table}

\subsection{Batch Effect Evaluation}

\begin{figure}[!htp]
    \centering
    \begin{subfigure}[b]{0.25\linewidth}
        \centering
        \includegraphics[width=1.1\linewidth]{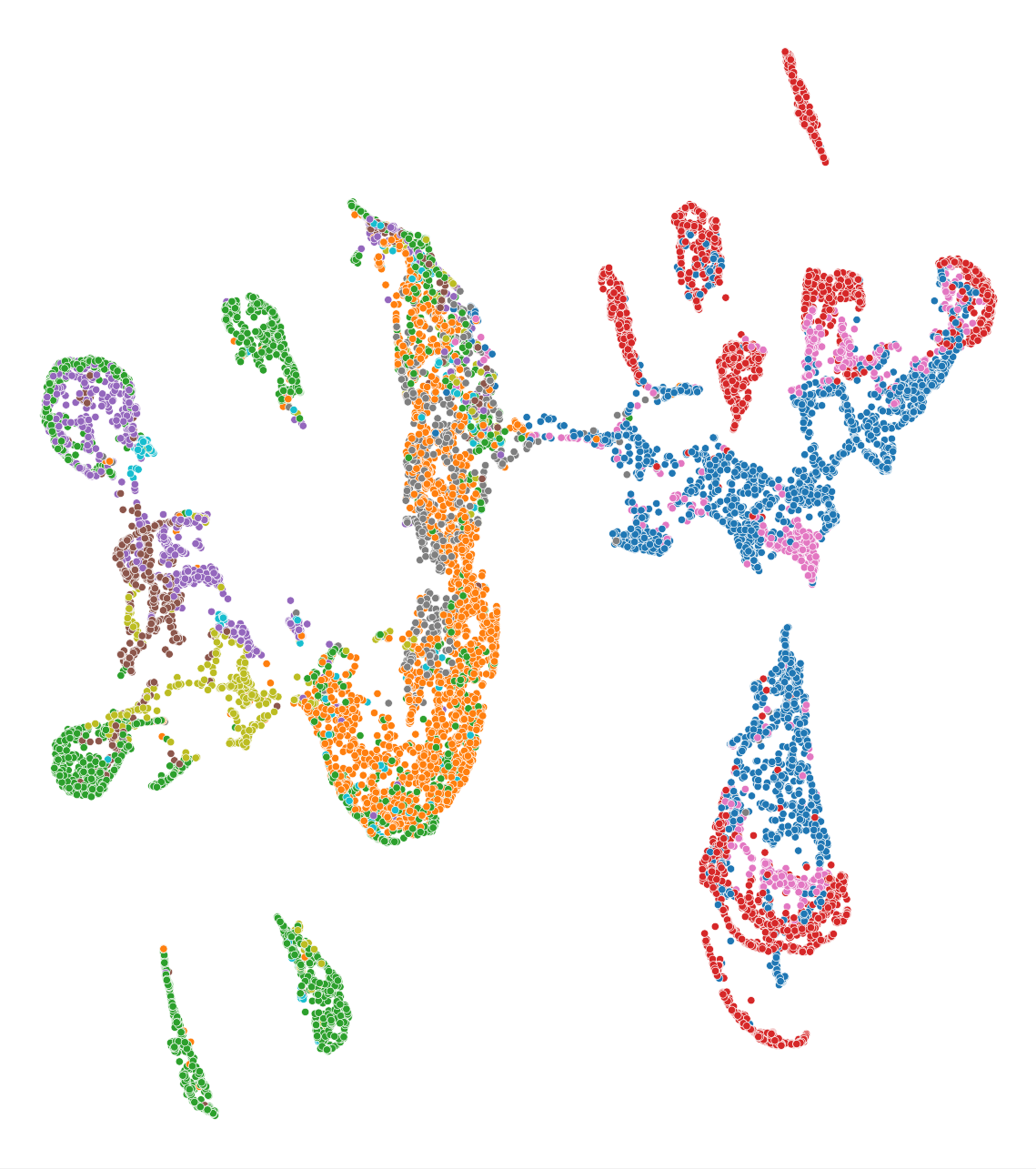}
        \caption{CLIP}
    \end{subfigure}\hspace{7mm}
    \begin{subfigure}[b]{0.25\linewidth}
        \centering
        \includegraphics[width=1.1\linewidth]{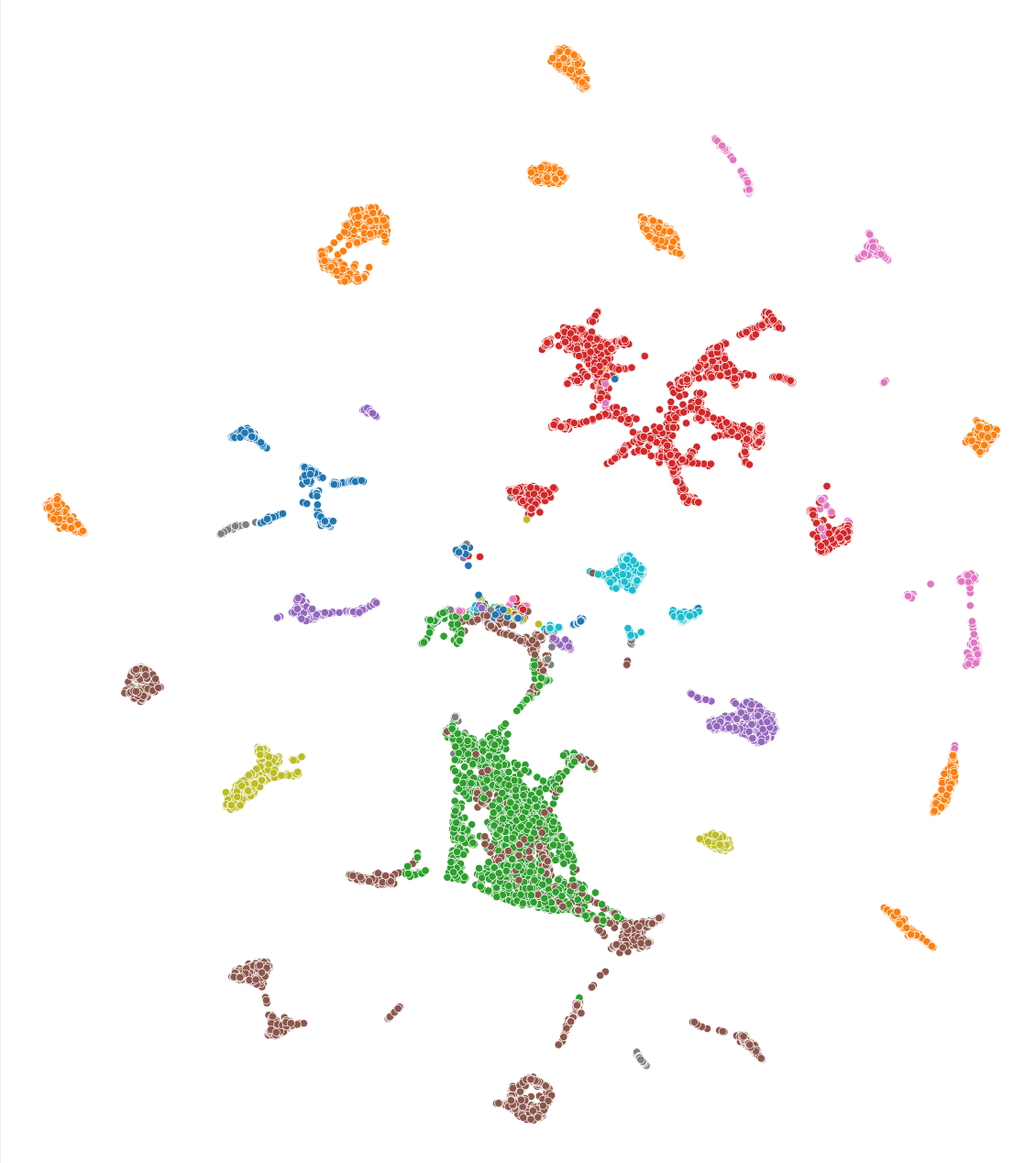}
        \caption{EMM}
    \end{subfigure}\hspace{7mm}
    \begin{subfigure}[b]{0.25\linewidth}
        \centering
        \includegraphics[width=1.1\linewidth]{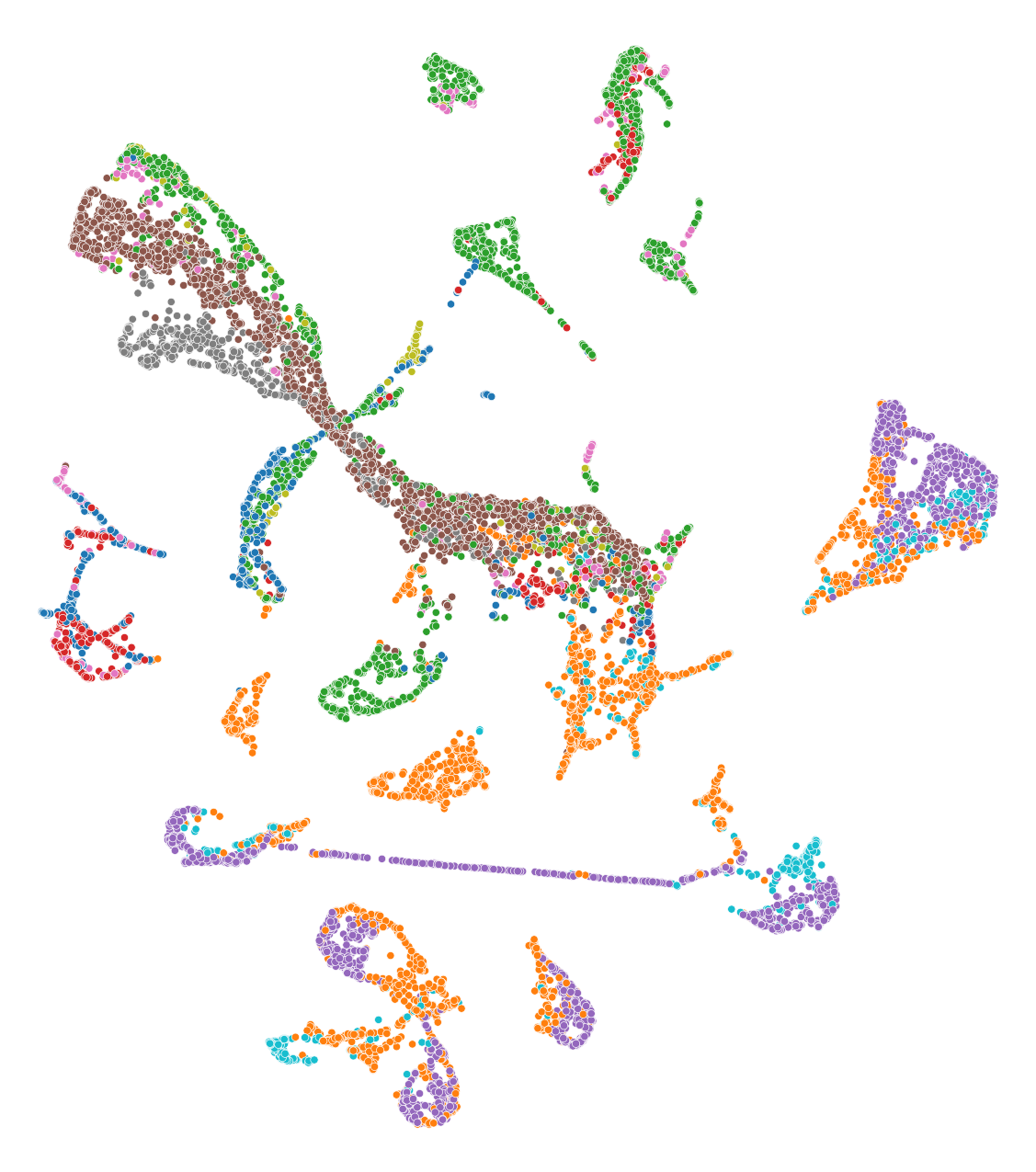}
        \caption{IMM}
    \end{subfigure}
    \caption{UMAP projections on the train set used for batch effect evaluation. Ideally, sources (colors) would be mixed in all the clusters.}
    \label{fig:big_be}
\end{figure}

The second evaluation focuses on batch effect. Table \ref{tab:batch_effect_results} encapsulates the results of a Logistic Regression classification. A notable observation is the pronounced improvement in generalization by our multiview loss compared to the baseline. Specifically, the generalization score jumps from 0.72 to 0.83 for EMM and a striking 0.93 for IMM. While source generalization emerges as particularly promising, the outcomes for plate and batch generalization are comparatively less definitive.
In order to provide a qualitative view of source batch effect, Figure \ref{fig:big_be} presents a UMAP projection of the image embeddings from the TARGET2 dataset, distinctly color-coded by source.

\begin{table}[h]
    \small
    \centering
    \caption{Results of the Batch Effect evaluation task on the main models using a Logistic Regression as the classifier on the frozen embeddings. $Acc_{Rand}$ denotes the accuracy on the random split, while $G_{NSP}$, $G_{NSB}$, and $G_{NSS}$ respectively denote the generalisation metric in the Plate, Source, and Batch separated splits.}
    \begin{tabular}{l|cccc}
        \hline
        Experiment & $Acc_{Rand} \uparrow$ & $G_{NSP} \uparrow$ & $G_{NSB} \uparrow$ & $G_{NSS} \uparrow$ \\
        \hline
        CLIP & 0.29 & 0.91 & 0.94 & 0.72 \\
        EMM & 0.36 & 0.92 & \textbf{0.95} & 0.83 \\
        IMM & 0.23 & \textbf{0.96} & 0.93 & \textbf{0.93} \\
        \hline
    \end{tabular}
    \label{tab:batch_effect_results}
\end{table}

\section{Conclusion}

In this paper, we propose two novel cross-modal pre-training objectives namely EMM and IMM. By introducing weak supervision that naturally occurs in HCS, our losses are able to learn better cross-modal representations and are more robust to batch effects. 
Future directions include designing specific encoders for cellular images and optimizing cross-modal fusion mechanisms.

\bibliographystyle{IEEEbib}
\bibliography{refs}

\begin{thebibliography}{10}

\bibitem{b1}
Karren~D. Yang, Anastasiya Belyaeva, Saradha Venkatachalapathy, Karthik Damodaran, Adityanarayanan Radhakrishnan, Abigail Katcoff, G.~V. Shivashankar, and Caroline Uhler,
\newblock ``Multi-domain translation between single-cell imaging and sequencing data using autoencoders,''
\newblock {\em Nature Communications}, vol. 12, 2019.

\bibitem{b3}
Yuyang Wang, Jianren Wang, Zhonglin Cao, and Amir~Barati Farimani,
\newblock ``Molecular contrastive learning of representations via graph neural networks,''
\newblock {\em Nature Machine Intelligence}, vol. 4, pp. 279 -- 287, 2021.

\bibitem{b4}
Srinivas~Niranj Chandrasekaran, Hugo Ceulemans, Justin~D. Boyd, and Anne~E Carpenter,
\newblock ``Image-based profiling for drug discovery: due for a machine-learning upgrade?,''
\newblock {\em Nature Reviews. Drug Discovery}, vol. 20, pp. 145 -- 159, 2020.

\bibitem{b8}
Mark-Anthony Bray, Shantanu Singh, Han Han, Chadwick~T. Davis, Blake Borgeson, Cathy~L. Hartland, Maria Kost-Alimova, Sigr{\'u}n~Margr{\'e}t G{\'u}stafsd{\'o}ttir, Christopher~C. Gibson, and Anne~E Carpenter,
\newblock ``Cell painting, a high-content image-based assay for morphological profiling using multiplexed fluorescent dyes,''
\newblock {\em Nature Protocols}, vol. 11, pp. 1757--1774, 2016.

\bibitem{b5}
Tim Becker, Kevin Yang, Juan~C. Caicedo, Bridget~K. Wagner, Vlado Danc{\'i}k, Paul~A. Clemons, Shantanu Singh, and Anne~E Carpenter,
\newblock ``Predicting compound activity from phenotypic profiles and chemical structures,''
\newblock {\em Nature Communications}, vol. 14, 2023.

\bibitem{b7}
Ethan Cohen, Maxime Corbe, Cl{\'a}udio~A Franco, Francisca~F Vasconcelos, Franck Perez, Elaine Del~Nery, Guillaume Bollot, and Auguste Genovesio,
\newblock ``Cell painting transfer increases screening hit rate,''
\newblock {\em Biological Imaging}, vol. 3, pp. e4, 2023.

\bibitem{b9}
Srinivas~Niranj Chandrasekaran, Jeanelle Ackerman, and Eric~Alix et~al,
\newblock ``Jump cell painting dataset: morphological impact of 136,000 chemical and genetic perturbations,''
\newblock {\em bioRxiv}, 2023.

\bibitem{b10}
Alec Radford, Jong~Wook Kim, and Chris~Hallacy et~al,
\newblock ``Learning transferable visual models from natural language supervision,''
\newblock in {\em International Conference on Machine Learning}, 2021.

\bibitem{b12}
Amanpreet Singh, Ronghang Hu, Vedanuj Goswami, Guillaume Couairon, Wojciech Galuba, Marcus Rohrbach, and Douwe Kiela,
\newblock ``Flava: A foundational language and vision alignment model,''
\newblock {\em 2022 IEEE/CVF Conference on Computer Vision and Pattern Recognition (CVPR)}, pp. 15617--15629, 2021.

\bibitem{b13}
Ye~Wang, Bo‐Shu Jiang, Changqing Zou, and Rui Ma,
\newblock ``Mxm-clr: A unified framework for contrastive learning of multifold cross-modal representations,''
\newblock {\em ArXiv}, vol. abs/2303.10839, 2023.

\bibitem{b14}
David Rogers and Mathew Hahn,
\newblock ``Extended-connectivity fingerprints,''
\newblock {\em Journal of chemical information and modeling}, vol. 50 5, pp. 742--54, 2010.

\bibitem{b16}
Thomas Kipf and Max Welling,
\newblock ``Semi-supervised classification with graph convolutional networks,''
\newblock {\em ArXiv}, vol. abs/1609.02907, 2016.

\bibitem{b18}
Xiyuan Wang and Muhan Zhang,
\newblock ``How powerful are spectral graph neural networks,''
\newblock in {\em International Conference on Machine Learning}, 2022.

\bibitem{b21}
Weihua Hu, Bowen Liu, Joseph Gomes, Marinka Zitnik, Percy Liang, Vijay~S. Pande, and Jure Leskovec,
\newblock ``Strategies for pre-training graph neural networks,''
\newblock {\em arXiv: Learning}, 2019.

\bibitem{b23}
Shuangjia Zheng, Jiahua Rao, Jixian Zhang, Chengtao Li, and Yuedong Yang,
\newblock ``Cross-modal graph contrastive learning with cellular images,''
\newblock {\em bioRxiv}, 2022.

\bibitem{b24}
Ana S{\'a}nchez-Fern{\'a}ndez, Elisabeth Rumetshofer, Sepp Hochreiter, and G{\`E}unter Klambauer,
\newblock ``C ontrastive learning of image - and structure based representations in drug discovery,''
\newblock 2022.

\bibitem{b25}
Cuong~Q. Nguyen, Dante~A. Pertusi, and Kim Branson,
\newblock ``Molecule-morphology contrastive pretraining for transferable molecular representation,''
\newblock {\em bioRxiv}, 2023.

\bibitem{b26}
Gabriele Corso, Luca Cavalleri, Dominique Beaini, Pietro Li{\`o}, and Petar Veli{\v{c}}kovi{\'c},
\newblock ``Principal neighbourhood aggregation for graph nets,''
\newblock {\em Advances in Neural Information Processing Systems}, vol. 33, pp. 13260--13271, 2020.

\bibitem{b28}
Hannes St{\"a}rk, Dominique Beaini, Gabriele Corso, Prudencio Tossou, Christian Dallago, Stephan G{\"u}nnemann, and Pietro Li{\`o},
\newblock ``3d infomax improves gnns for molecular property prediction,''
\newblock in {\em International Conference on Machine Learning}. PMLR, 2022, pp. 20479--20502.

\bibitem{b29}
Maciej Sypetkowski, Morteza Rezanejad, Saberian, et~al.,
\newblock ``Rxrx1: A dataset for evaluating experimental batch correction methods,''
\newblock in {\em Proceedings of the IEEE/CVF Conference on Computer Vision and Pattern Recognition}, 2023, pp. 4284--4293.

\bibitem{b30}
Ihab Bendidi, Adrien Bardes, Ethan Cohen, Alexis Lamiable, Guillaume Bollot, and Auguste Genovesio,
\newblock ``No free lunch in self supervised representation learning,''
\newblock {\em arXiv preprint arXiv:2304.11718}, 2023.

\end{thebibliography}

\end{document}